\documentclass[letterpaper,10pt,conference]{ieeeconf}
\IEEEoverridecommandlockouts
\usepackage{amsmath,amssymb,graphicx,booktabs,enumerate}
\usepackage{cite}
\usepackage[hidelinks]{hyperref}
\hypersetup{pdftitle={Safety-Constrained Intelligent Control of a 2-DOF Robotic Manipulator Using Control Barrier Functions}}
\newcommand{\R}{\mathbb{R}}
\newcommand{\norm}[1]{\left\|#1\right\|}
\newcommand{\dd}{\mathrm{d}}
\title{Safety-Constrained Intelligent Control of a 2-DOF Robotic\\Manipulator Using Control Barrier Functions}
\author{Gabriel da Silva Lima$^{1}$, Made Widhi Surya Atman$^{2}$, and Wallace Moreira Bessa$^{1}$
\thanks{$^{1}$G. S. Lima and W. M. Bessa are with the Smart Systems Lab, University of Turku, 20520 Turku, Finland (e-mail: gdasil@utu.fi, wmobes@utu.fi).}%
\thanks{$^{2}$M. W. S. Atman is with the Department of Computing, University of Turku, 20520 Turku, Finland (e-mail: widhi.atman@utu.fi).}} 
\begin{document}
\maketitle
\thispagestyle{empty}
\pagestyle{empty}

\begin{abstract}
Control barrier function (CBF) filters depend on dynamic models that can be inaccurate when their parameters and external forces are uncertain. This paper investigates a radial basis function (RBF) neural network compensator whose output is shared by a Cartesian tracking controller and a joint-space CBF filter. An auxiliary state driven by the difference between requested and applied torques corrects the learning signal, reducing the direct influence of filter intervention on adaptation. A two-link manipulator is evaluated through constrained regulation, ablations, matched uncertainty and disturbance trials, and operating-range and timing studies. The results demonstrate that the proposed controller improves tracking while maintaining joint, velocity, and torque limits under parameter uncertainty, and that both the shared compensation and the auxiliary correction contribute to this improvement.
\end{abstract}

\section{Introduction}
Robotic manipulators must track motions within joint, velocity, and actuator limits. Model-based control tracks accurately, but inertia errors and unmodeled forces corrupt both the tracking law and the constraint prediction. A control barrier function (CBF) filter offers a convenient way to modify a requested input through a quadratic program (QP)~\cite{ames2017}. Position constraints require a higher-relative-degree construction, while joint-velocity constraints depend directly on acceleration~\cite{xiao2022}. Both constructions remain sensitive to errors in the assumed dynamics.

Neural compensation has been demonstrated in intelligent controllers. Lima~\emph{et al.}\ combine sliding mode  control and online neural adaptation for a single-link flexible manipulator~\cite{lima2021}. Alwala~\emph{et al.}\ integrate an intelligent controller with an RBF neural compensator with EKF-based multisensor fusion for differential-drive robots with unmodeled dynamics~\cite{alwala2026}. These works show that online neural compensation is effective.

CBFs also support robot navigation and coordination: Keyumarsi~\emph{et al.}\ synthesize barrier functions online from LiDAR data using Gaussian processes~\cite{keyumarsi2024}, and Pham~\emph{et al.}\ combine distributed QPs and CBFs for formation control with rigid and elastic edges and obstacle avoidance~\cite{pham2024}. Here, the joint-limit geometry is known; uncertainty affects the dynamics used to predict constraint evolution.

Robust CBFs and input-to-state safety tolerate bounded model error through constraint margins~\cite{jankovic2018,kolathaya2019}. Adaptive CBFs couple parameter estimation and safety conditions~\cite{taylor2020,lopez2021}, and learning-based approaches estimate the residual that enters the barrier condition~\cite{taylor2020learning}. Wang~\emph{et al.}\ combine neural estimation with augmented high-order CBFs and auxiliary systems that adjust barrier conditions for constrained robotic tracking~\cite{wang2025}. Sweatland~\emph{et al.}\ use adaptive neural identification and an estimation-error bound to formulate CBF constraints~\cite{sweatland2026}. We examine how an online neural network compensator interacts with a filter when both use an incomplete model.

Two issues arise when the controller and the filter share an incomplete model. First, compensating only the tracking controller leaves the filter unaware of the learned correction, so the two components may request incompatible accelerations. Second, a tracking-driven adaptation law can react to error generated by deliberate filter intervention. We investigate a shared RBF correction and an auxiliary state that accounts for the nominal Cartesian effect of the torque modification. Although pseudo-control hedging can remove the effect of actuator limits from the signal seen by an adaptive law~\cite{johnson2003}, compensated tracking errors can play the same role in command-filtered backstepping~\cite{farrell2009}, and auxiliary systems driven by the saturation error are standard in adaptive neural control under input constraints~\cite{chen2011}, these designs treat the gap between commanded and applied inputs as an intentional deviation that should not be learned. In this work, we apply this principle to CBF intervention, which, unlike saturation, depends on the state, the barrier rates, and the learned term itself. The auxiliary state corrects only the adaptation signal; it does not adjust the barrier functions or restore feasibility.

Overral, the main contributions of this paper are as follows:
\begin{enumerate}[i.]
 \item We adapt the learning input signal to CBF intervention: an auxiliary
 state removes the nominal effect of the filter's torque modification from
 the signal used to train an RBF compensator shared by the tracking
 controller and the CBF filter.
 \item We relate the corrected signal to the joint-acceleration residual that
 the filter must bound, and show that this residual is driven mainly by the
 transient error of the shared correction rather than by model uncertainty.
 \item We isolate the effects of sharing the RBF output and of correcting its
 learning signal through matched ablations under fixed tuning.
 \item We evaluate tracking performance, observed constraint satisfaction,
 residual-bound validity, QP feasibility, and computation time as separate
 outcomes, over 172 simulations including uncertainty, disturbance, and
 operating-range studies.
\end{enumerate}


\section{Manipulator Model and Control Objective}
Let $q=[q_1,q_2]^\top$, $v=\dot q$, and $\tau\in\R^2$ denote joint position, velocity, and torque. The planar manipulator satisfies
\begin{equation}
 \dot q=v,\qquad M(q)\dot v=F(q,v)+\tau+d_\tau(t,q,v).
 \label{eq:plant}
\end{equation}
Gravity is absent from the implemented dynamics. With link lengths $l_i$, center-of-mass distances $r_i$, masses $m_i$, and inertias $I_i$, its inertia matrix is
\begin{equation}
\begin{aligned}
 M_{11}&=m_1r_1^2+m_2(l_1^2+r_2^2+2l_1r_2\cos q_2)+I_1+I_2,\\
 M_{12}&=M_{21}=m_2(r_2^2+l_1r_2\cos q_2)+I_2,\\
 M_{22}&=m_2r_2^2+I_2.
\end{aligned}\label{eq:inertia}
\end{equation}
Writing $b=m_2l_1r_2$, the implemented velocity-dependent term and unmodeled forces are
\begin{equation}
\begin{aligned}
 F&=b\sin q_2[2v_1v_2+v_2^2,\;-v_1^2]^\top,\\
 d_\tau&=[-b_1v_1,\;-b_2(v_2-v_1)]^\top+J(q)^\top f_{\rm ext}(t).
\end{aligned}\label{eq:forces}
\end{equation}
Here $J=\partial p/\partial q$ is the Jacobian of
\begin{equation}
 p(q)=\begin{bmatrix}l_1\cos q_1+l_2\cos(q_1+q_2)\\
 l_1\sin q_1+l_2\sin(q_1+q_2)\end{bmatrix}.
\end{equation}
Kinematic parameters and joint states are available to the controller. The estimated model uses $\hat M$ and $\hat F$ with uncertain masses and inertias and omits $d_\tau$. 

The objective is regulation to a feasible constant Cartesian target $p_d$ subject to componentwise bounds
\begin{equation}
 q_i^-\le q_i\le q_i^+,\quad v_i^-\le v_i\le v_i^+,\quad |\tau_i|\le\bar\tau.
 \label{eq:limits}
\end{equation}
Define $e=p-p_d$, $s=\dot e+\Lambda e$, and $\Lambda\succ0$ diagonal. All experiments use a fixed target. Singularity avoidance is not implemented; the pseudoinverse is used numerically, and the cancellation below assumes nonsingular $J$.

\section{Shared Neural Compensation and Barrier Filtering}

\subsection{Control law}
Let $\hat d\in\R^2$ be a Cartesian acceleration correction. The requested torque is
\begin{equation}
 u=-\hat F+\hat M J^\dagger
 (\ddot p_d-\dot Jv-\Lambda\dot e-\Lambda s-\hat d).
 \label{eq:nominal}
\end{equation}
The filter returns $\tau$, and $\Delta\tau=\tau-u$ denotes its intervention. With $\hat G=J\hat M^{-1}$, introduce the continuous-time auxiliary interpretation
\begin{equation}
 \dot z=-\Lambda z+\hat G\Delta\tau,\qquad s_c=s-z,\qquad z(0)=0.
 \label{eq:aux}
\end{equation}
To explain this choice without assuming that the neural estimate is exact, define the actual Cartesian model mismatch at the applied torque by
\begin{equation}
 \Delta_p(\tau)=J\!\left[M^{-1}(F+\tau+d_\tau)
                -\hat M^{-1}(\hat F+\tau)\right].
 \label{eq:mismatch}
\end{equation}
For nonsingular $J$, substitution into the plant gives
\begin{equation}
\begin{aligned}
 \dot s&=-\Lambda s-\hat d+\hat G\Delta\tau+\Delta_p(\tau),\\
 \dot s_c&=-\Lambda s_c+\Delta_p(\tau)-\hat d.
\end{aligned}\label{eq:sc}
\end{equation}
Thus, the explicit nominal effect of the filter disappears from the corrected signal dynamics. 

Following~\cite{lima2021}, the RBF network for each Cartesian component takes the corrected signal as input and uses Gaussian basis functions:
\begin{equation}
 a_{ij}(s_{c,i})=\exp\!\left[-\frac{(s_{c,i}-c_j)^2}{2\ell_j^2}\right],
 \qquad \hat d_i=w_i^\top a_i.
 \label{eq:rbf}
\end{equation}
The learning rule for the RBF weights is defined as follows~\cite{lima2021}:
\begin{equation}
 \dot{w}_{i} = \mathcal P_\mu(w_{i},g_{i}),\label{eq:weights}
\end{equation}
where $g_{i}=\eta_i s_{c,i}a_i$ and the original projection-style map is retained. Let $\mathcal A(w,g)$ denote $\norm w<\mu$ or the conjunction $\norm w=\mu$ and $w^\top g<0$. Then
\begin{equation}
\mathcal P_\mu(w,g)=
\begin{cases}
g,&\text{if }\mathcal A(w,g),\\
\left(I-\dfrac{ww^\top}{w^\top w}\right)g,&\text{otherwise}.
\end{cases}\label{eq:projection}
\end{equation}

 \label{eq:zdiscrete}

\subsection{CBF inequalities with an incomplete model}
The same learned output is mapped into the joint-acceleration prediction:
\begin{equation}
 \hat a(\tau)=\hat f+\hat B\tau,\quad
 \hat B=\hat M^{-1},\quad
 \hat f=\hat B\hat F+J^\dagger\hat d.
 \label{eq:prediction}
\end{equation}
Let $r=\dot v-\hat a(\tau)$ be the acceleration residual. Constants $\rho_i>0$ represent the assumed componentwise bounds $|r_i|\le\rho_i$. For $h_i^-=q_i-q_i^-$ and $h_i^+=q_i^+-q_i$, an exponential (second-order) barrier construction~\cite{nguyen2016,xiao2022} with positive rates $\alpha_i,\beta_i$, tightened by $\rho_i$ as in robust CBFs~\cite{jankovic2018}, yields
\begin{equation}
\begin{aligned}
 \hat a_i-\rho_i+(\alpha_i+\beta_i)v_i
                     +\alpha_i\beta_i h_i^-&\ge0,\\
 -\hat a_i-\rho_i-(\alpha_i+\beta_i)v_i
                     +\alpha_i\beta_i h_i^+&\ge0.
\end{aligned}\label{eq:positioncbf}
\end{equation}
The first-order velocity conditions, with rates $k_i^-,k_i^+>0$, are
\begin{equation}
\begin{aligned}
 \hat a_i-\rho_i+k_i^-(v_i-v_i^-)&\ge0,\\
 -\hat a_i-\rho_i+k_i^+(v_i^+-v_i)&\ge0.
\end{aligned}\label{eq:velocitycbf}
\end{equation}
The applied torque solves the strictly convex two-variable QP
\begin{equation}
\begin{aligned}
 \tau=\arg\min_{\xi\in\R^2}\quad&\tfrac12(\xi-u)^\top W(\xi-u)\\
 \text{subject to}\quad&\eqref{eq:positioncbf},\ \eqref{eq:velocitycbf}
                  \text{ evaluated at }\hat a(\xi),\\
 &-\bar\tau\mathbf1\le\xi\le\bar\tau\mathbf1,
\end{aligned}\label{eq:qp}
\end{equation}
with $W=I$. There are no slack variables or fallback inputs; a failed QP terminates the numerical run.

\subsection{Conditional safety interpretation}
If the residual bound holds continuously at the applied input, the constraints above imply their corresponding inequalities for the true acceleration. For example, $\ddot h_i^-+(\alpha_i+\beta_i)\dot h_i^-+\alpha_i\beta_i h_i^-\ge0$. Defining $\psi_i^-=\dot h_i^-+\alpha_i h_i^-$ gives $\dot\psi_i^-+\beta_i\psi_i^-\ge0$. Nonnegative initial values of $h_i^-$ and $\psi_i^-$ then imply $h_i^-(t)\ge0$ by successive comparison arguments. The upper-position and velocity cases follow analogously. This standard argument~\cite{xiao2022} also requires existing solutions, feasibility, and admissible initial values of $\psi_i^\pm$.

These conditions are not established by the present implementation. The QP is solved every $T_c$, its torque is held between updates, and no intersample robustness margin is included. Sampled-data invariance needs additional conditions~\cite{breeden2022}. Furthermore, an empirical residual diagnostic
\begin{equation}
 R=\max_{k,i}\frac{|r_i(t_k)|}{\rho_i}\label{eq:R}
\end{equation}
can falsify an assumed bound when $R>1$, but $R\le1$ on one trajectory cannot certify it uniformly. We therefore report task error, observed constraint satisfaction, and residual-bound satisfaction as distinct outcomes.

At a nonsingular configuration, from~\eqref{eq:mismatch} and~\eqref{eq:prediction} we can calculate the residual $r$ by
\begin{equation}
 r=J^{-1}\bigl(\Delta_p(\tau)-\hat d\bigr).
 \label{eq:residualmap}
\end{equation}



\subsection{Lyapunov stability analysis}\label{sec:lyapunov}
Consider Equation~\eqref{eq:weights}, with $w_i(0)=0$, and let $\lambda_i>0$ be the $i$th diagonal entry of $\Lambda$.
Assume that, on a compact set $\Omega$ containing the closed-loop trajectory, $J$ is nonsingular, the QP~\eqref{eq:qp} is feasible, and there exist $w_i^\star$ with $\norm{w_i^\star}\le\mu$ and $\bar\varepsilon_i\ge0$ such that
\begin{equation}
 \Delta_{p,i}=w_i^{\star\top}a_i(s_{c,i})+\varepsilon_i,\qquad
 |\varepsilon_i|\le\bar\varepsilon_i.
 \label{eq:ideal}
\end{equation}
With $\tilde w_i=w_i^\star-w_i$, the $i$th row of~\eqref{eq:sc} becomes $\dot s_{c,i}=-\lambda_i s_{c,i}+\tilde w_i^\top a_i+\varepsilon_i$. Consider
\begin{equation}
 V_i=\tfrac12 s_{c,i}^2+\tfrac{1}{2\eta_i}\tilde w_i^\top\tilde w_i .
\end{equation}
Since $\dot{\tilde w}_i=-\dot w_i$, the terms $\pm s_{c,i}\tilde w_i^\top a_i$ cancel and
\begin{equation}
 \dot V_i=-\lambda_i s_{c,i}^2+s_{c,i}\varepsilon_i
 -\tfrac{1}{\eta_i}\tilde w_i^\top\bigl(\mathcal P_\mu(w_i,g_i)-g_i\bigr).
\end{equation}
The last term vanishes when $\mathcal A(w_i,g_i)$ holds. Otherwise $\norm{w_i}=\mu$ and $w_i^\top g_i\ge0$, so $\tilde w_i^\top(\mathcal P_\mu-g_i) =-(\tilde w_i^\top w_i)(w_i^\top g_i)/\mu^2\ge0$, because $\tilde w_i^\top w_i\le\mu\norm{w_i^\star}-\mu^2\le0$. Projection also keeps $\norm{w_i}\le\mu$, so $\norm{\tilde w_i}\le2\mu$. Hence
\begin{equation}
 \dot V_i\le-\lambda_i s_{c,i}^2+\bar\varepsilon_i|s_{c,i}| \le-\tfrac{\lambda_i}{2}s_{c,i}^2+\tfrac{\bar\varepsilon_i^2}{2\lambda_i}.
 \label{eq:vdot}
\end{equation}
Thus $s_{c,i}$ and $w_i$ remain bounded, and integrating~\eqref{eq:vdot} gives
\begin{equation}
 \frac1T\int_0^T s_{c,i}^2\,\dd t\le\frac{2V_i(0)}{\lambda_i T} +\frac{\bar\varepsilon_i^2}{\lambda_i^2},
\end{equation}
so the mean-square corrected error is ultimately bounded by $(\bar\varepsilon_i/\lambda_i)^2$. If $\bar\varepsilon_i=0$, then $s_{c,i}\in\mathcal L_2\cap\mathcal L_\infty$ with $\dot s_{c,i}$ bounded on $\Omega$, and Barbalat's lemma gives $s_{c,i}\to0$.


\section{Numerical Protocol}
The physical parameters are $l=(0.35,0.25)$ m, $r=(0.175,0.125)$ m, $m=(3,2)$ kg, and $I_i=m_ir_i^2/12$, as implemented. The default estimated masses and inertias are $80\%$ of these values. Unmodeled force coefficients are $b_1=0.12$ and $b_2=0.10$ N\,m\,s/rad. Initially, $q=(0.1,-1.3)$ rad, $v=0$, and all weights are zero. The target is $p_d=p(1.4,-1.3)=(0.30824,0.36987)$ m.

Joint limits are $q_1\in[0,\pi/2]$, $q_2\in[-\pi/2,\pi/2]$, $|v_1|\le0.5$ rad/s, and $|v_2|\le0.2$ rad/s. Unless stated otherwise, $\bar\tau=0.5$ N\,m, $\rho_1=1.25$ and $\rho_2=1$ rad/s$^2$ (see Section~\ref{sec:block3}), $\Lambda=0.5I$ s$^{-1}$, $\alpha_i=\beta_i=4$ s$^{-1}$, and $k_i^-=k_i^+=6$ s$^{-1}$. For both RBFs, $\eta_i=10$, $\mu=100$, and
\begin{equation*}
\begin{aligned}
 c&=0.1[-1,-1/2,-1/4,0,1/4,1/2,1],\\
 \ell&=0.1[1/2,1/3,1/4,1/8,1/4,1/3,1/2],
\end{aligned}
\end{equation*}
with centers and widths in the units of $s_c$ (m/s). Gains are fixed across methods and trials.

The controller period is $T_c=0.02$ s, with fourth-order Runge--Kutta integration at $0.005$ s and held torques. The inclusive control grid runs from 0 through 30 s, producing 1501 controls and a final state at $T=30.02$ s. 

State limits are checked at every integration endpoint and input limits at control instants, with reporting tolerance $10^{-7}$. The residual is evaluated offline at accepted control instants using the true plant at the same state and applied torque. Cartesian RMSE is $[T^{-1}\int_0^T\norm{e(t)}^2\dd t]^{1/2}$, numerically integrated by the trapezoidal rule. Settling requires $\norm e\le1$ mm at every subsequent saved state. Failed runs remain in all completion denominators; their partial-horizon errors are excluded from full-horizon comparisons.

\section{Results}

Four blocks of simulations were conducted. Block~1 evaluates constrained regulation of the complete scheme in a reference case and under a tighter torque limit. The second group isolates the role of each component through ablations that remove the neural compensation, restrict it to the tracking controller, or remove the auxiliary correction. The third block tests robustness to randomized inertial and damping parameters and to external force pulses applied to the end effector. Block~4 explores the operating range over torque limits and model errors, and measures the computation time of the controller.

\subsection{Constrained regulation}
\begin{figure}[t]
\centering\includegraphics[width=0.92\columnwidth]{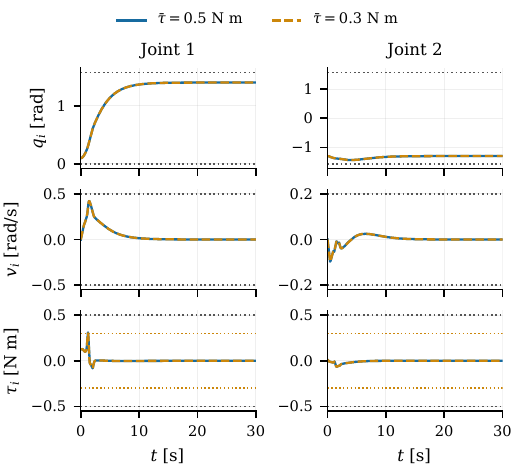}
\caption{Block 1: complete scheme with torque limits of 0.5 and 0.3 N\,m. Dotted lines denote the corresponding bounds. Joint position and velocity use every saved integration endpoint; torque uses control instants.}
\label{fig:block1}
\end{figure}

Figure~\ref{fig:block1} compares the reference torque limit with a tightened limit of $0.3$ N\,m. Both runs complete with all 1501 QPs reported optimal and no sampled joint, velocity, or torque violation. At $0.5$ N\,m, the Cartesian RMSE is $0.15955$ m, the final error is $2.699\times10^{-6}$ m, and settling occurs at $17.075$ s. The peak applied torques are $0.3090$ and $0.0655$ N\,m. Tightening the limit clips the first-joint peak to $0.3$ N\,m at three control instants, with RMSE $0.15957$ m and the same settling time.

The residual ratio is $R=0.8691$ in the reference case and $0.8686$ with the tightened limit, so the assumed bound holds at every accepted control instant of these trajectories, which does not certify it elsewhere. The reference weight norms remain below $0.563$, well inside $\mu=100$, and the minimum recorded Jacobian singular value is $0.1629$ m/rad. 

\subsection{Component ablations}
Four methods share initial conditions, bounds, gains, and estimated dynamics. Method A disables learning and keeps $\hat d=0$. Method B uses the RBF and corrected learning signal in the tracking controller but removes $J^\dagger\hat d$ from the CBF prediction. Method C shares the RBF output but replaces $s_c$ by $s$ in both the basis input and adaptation update. Method D is the complete scheme. Consequently, C tests removal of the whole auxiliary correction, rather than isolating only a change in the weight-update multiplier.

\begin{table}[t]
\centering
\caption{Block 2: deterministic comparison at fixed tuning}
\label{tab:ablation}
\begin{tabular}{@{}lrrrr@{}}
\toprule
Method & End [s] & RMSE [m] & Final error [m] & $R$\\
\midrule
A: no RBF &30.02&0.25423&0.27499&0.538\\
B: controller only &30.02&0.19351&0.09480&0.937\\
C: no correction &1.14&---&---&2.229\\
D: complete &30.02&0.15955&$2.70\!\times\!10^{-6}$&0.869\\
\bottomrule
\end{tabular}
\end{table}

\begin{figure}[t]
\centering\includegraphics[width=0.92\columnwidth]{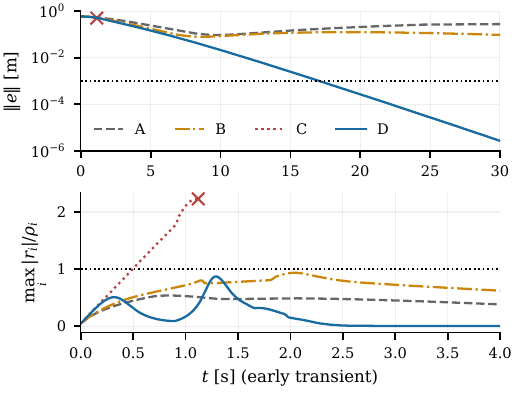}
\caption{Block 2: Cartesian error and early acceleration-residual ratio for methods A--D. The dotted error threshold is 1 mm; the residual threshold is one. Crosses mark the ends of C's recorded traces.}
\label{fig:block2}
\end{figure}

Table~\ref{tab:ablation} and Fig.~\ref{fig:block2} show that A respects sampled constraints and the residual bound but does not reach the target: it ends at $q_1\approx1.49$ rad, at the margin-tightened bound $\pi/2-\rho_1/(\alpha_1\beta_1)$, where the filter cancels the requested torque. B reduces error yet leaves a final offset of $9.48$ cm, with the lower joint-2 velocity condition active at 97\% of control instants. Such boundary stalling is a known property of CBF-QP filters~\cite{reis2021}. D reduces full-horizon RMSE by $37.2\%$ relative to A and $17.5\%$ relative to B, while reaching micrometre-scale error in this deterministic simulation. This supports sharing the learned correction under the tested tuning, not against separately optimized controllers.

C first violates a velocity limit at $0.68$ s (peak second-joint excess $0.1076$ rad/s) and terminates with an infeasible QP at $1.14$ s. The result is consistent with filter intervention contaminating the uncorrected learning signal, but does not prove that every uncorrected adaptive design fails.

\begin{figure}[!tb]
\centering\includegraphics[width=0.92\columnwidth]{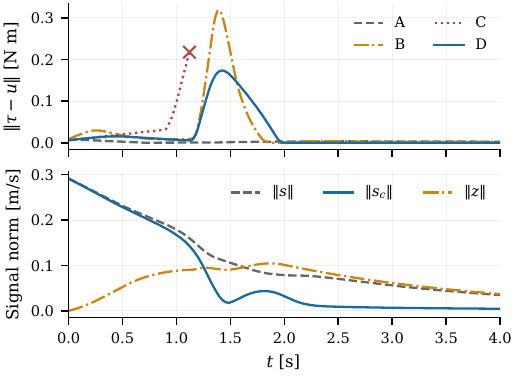}
\caption{Block 2: torque-intervention magnitude for A--D and the complete scheme's uncorrected, corrected, and auxiliary signals during the early transient. The cross marks C's last accepted control. The auxiliary correction changes the signal driving the RBF during filter intervention.}
\label{fig:auxiliary}
\end{figure}

\begin{figure*}[t]
\centering\includegraphics[width=\textwidth]{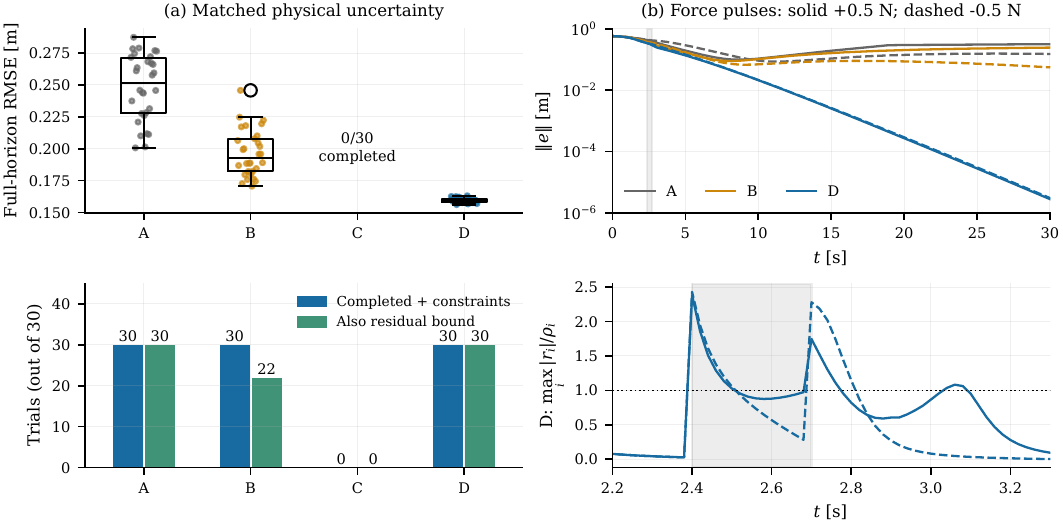}
\caption{Block 3: (a) full-horizon RMSE for completed trials and counts over all 30 matched realizations; C has no completed trials. ``Also residual bound'' requires completion, observed constraint satisfaction, and $R\le1$. (b) Both external-force signs, applied during the shaded interval $[2.4,2.7)$ s. Colors identify A, B, and D; solid/dashed lines identify positive/negative pulses. C stops before the pulse and is excluded from recovery curves. The lower-right panel shows D's residual ratios.}
\label{fig:block3}
\end{figure*}

Figure~\ref{fig:auxiliary} shows the mechanism: during intervention, $z$ separates the adaptation signal from the tracking signal through~\eqref{eq:aux}, without measuring acceleration. D's integrated intervention magnitude is $0.0940$ N\,m\,s.

\subsection{Uncertainty and external-force trials}\label{sec:block3}
Thirty physical realizations independently scale each of $m_1,m_2,I_1,I_2$ uniformly over $[0.85,1.15]$ times its reference value, and each damping coefficient over $[0.5,1.5]$ times its reference value. Every realization is applied to all four methods with the original estimated model, seed 20261103, and no retuning.

A, B, and D complete all 30 trials without observed constraint violations; C completes none. Median full-horizon RMSE values are $0.25179$, $0.19241$, and $0.15954$ m for A, B, and D, respectively (Fig.~\ref{fig:block3}). The complete scheme therefore preserves its tracking advantage across these matched perturbations. All 30 D trials also satisfy the residual bound throughout, with $R$ between $0.82$ and $0.97$, as do all 30 A trials and 22 B trials.

This outcome depends on a calibrated margin. With $\rho_1=1$ rad/s$^2$, the same 30 trials kept the bound in only five cases. Every exceedance was a joint-1 event of three to eight control instants within $[1.02,1.40]$ s, with $R\le1.20$, at the onset of velocity-limit intervention. There, the shared term $-(J^{\dagger}\hat d)_1$ contributed $1.44$--$1.83$ rad/s$^2$, whereas the input-map and unmodeled-force terms of~\eqref{eq:residualmap} summed to only $-0.39$ to $-0.80$ rad/s$^2$. The correction accumulated during acceleration overshoots the physical mismatch when the filter starts to decelerate joint 1, and $\norm{J^{-1}}\approx5.9$ rad/m amplifies the discrepancy as in~\eqref{eq:residualmap}. We therefore set $\rho_1=1.25$ rad/s$^2$, well below the joint-1 compatibility limit $k_1(v_1^+-v_1^-)/2=3$ rad/s$^2$. Because this value was chosen after observing those trials, $R\le1$ here is a consistency check rather than an independent validation. A constant margin cannot absorb all errors: the two joint-2 velocity conditions are compatible only if $\rho_2\le k_2(v_2^+-v_2^-)/2=1.2$ rad/s$^2$.

Eight further runs apply $f_{\rm ext}=[0,\pm0.5]^\top$ N during $2.4\le t<2.7$ s, one run per sign and method. D completes both cases without observed constraint violations, with final errors $3.02$ and $2.79\,\mu$m for negative and positive pulses. A violates sampled constraints for both signs; B does so for the positive sign. B's final errors are $0.0556$ and $0.2458$ m, respectively. C stops before the force starts. D's peak residual ratios, on joint 2, reach $2.432$ and $2.386$. The pulses thus show recovery while contradicting the residual-bound assumption during the disturbance; no time-step convergence study of the force discontinuities is claimed.

\subsection{Operating range and computation time}
\begin{figure}[!tb]
\centering\includegraphics[width=0.92\columnwidth]{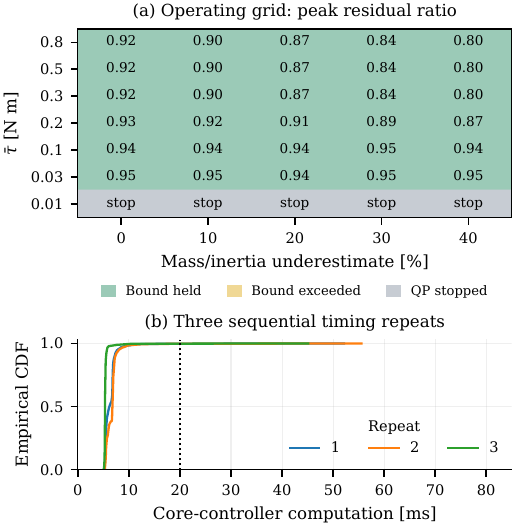}
\caption{Block 4: (a) exploratory operating grid for D. All completed cells respect sampled constraints and the residual bound; numeric entries give $R$. Stopped cells are not successful operating points. (b) empirical distributions of core computation time from three isolated sequential runs; the dotted line is the 20 ms control period.}
\label{fig:block4}
\end{figure}

A grid of 35 complete-method runs combines torque limits $\bar\tau\in\{0.01,0.03,0.1,0.2,0.3,0.5,0.8\}$ N\,m with common estimated mass/inertia underestimates of $0$, $10$, $20$, $30$, and $40\%$. The physical plant is fixed. All 30 cases with $\bar\tau\ge0.03$ N\,m complete without observed constraint violations and keep the assumed residual bound, with $R$ between $0.80$ and $0.95$. All five cases at $0.01$ N\,m stop with an infeasible QP, before any sampled state violation. Completion means reaching the horizon, not reaching the target (Fig.~\ref{fig:block4}).

The $0.01$ and $0.03$ N\,m rows were added after the initial $0.1$--$0.8$ N\,m grid produced no failures. This is an exploratory map, not a certified boundary. QP infeasibility reflects incompatible barrier and actuator inequalities at the encountered state, which also depend on the neural term and fixed margins, not physical impossibility of the task.

Three additional timing runs execute sequentially after all other experiments. Median core computation times are $6.17$, $6.77$, and $5.32$ ms; the corresponding 95th percentiles are $7.83$, $8.07$, and $5.71$ ms. Nevertheless, maximum times are $52.15$, $55.57$, and $45.11$ ms, and $4$, $5$, and $4$ of 1501 updates exceed 20 ms. The measurement covers kinematics, adaptation, nominal torque, and QP construction and solution, but not the auxiliary update, diagnostics, integration, or output. These are typical host costs, not a real-time guarantee.

\section{Concluding remarks}
In this paper, we proposed a safety-constrained intelligent controller for a two-link robotic manipulator that combines a Cartesian tracking law, an RBF neural compensator, and a joint-space CBF filter. The neural compensation is shared by the controller and the filter, so that both act on the same estimate of the unmodeled dynamics, and an auxiliary state removes the nominal effect of filter intervention from the learning signal, so that the network does not learn deviations deliberately introduced by the filter. We also related the corrected signal to the joint-acceleration residual that the filter must bound, and showed that this residual is driven mainly by the transient error of the shared correction.

The results demonstrated that the proposed controller improves tracking while maintaining joint, velocity, and torque limits under parameter uncertainty and external disturbances. The ablations showed that both the shared compensation and the auxiliary correction contribute to this improvement: without the auxiliary correction, the filter becomes infeasible early in every trial, and without sharing, a regulation offset remains. The analysis of the acceleration residual also clarified when the robust barrier conditions hold and how the margins should be chosen.

For future work, we aim to derive a residual bound that accounts for the transient error of the shared correction, so that the barrier margins can be selected a priori rather than calibrated. We also plan to incorporate sampled-data safety margins, a feasibility-preserving fallback for tight actuation, and singularity avoidance, and to validate the approach experimentally on a physical manipulator.

\section*{ACKNOWLEDGMENT}

We acknowledge the financial support of the Finnish Ministry of Education and Culture through the Digital Waters (DIWA) Doctoral Education Pilot, Decision No. VN/3137/2024-OKM-6, and through the Intelligent Work Machines Doctoral Education Pilot Program (IWM VN/3137/2024-OKM-4).

\bibliographystyle{IEEEtran}
\bibliography{references}

\end{document}